\documentclass[11pt]{article}

\usepackage{acl}

\usepackage{times}
\usepackage{latexsym}

\usepackage[T1]{fontenc}

\usepackage[utf8]{inputenc}

\usepackage{microtype}
\usepackage{comment}

\usepackage{inconsolata}

\usepackage{graphicx}

\usepackage{amsmath}
\usepackage{amssymb}
\usepackage{bbm}  
\usepackage{graphicx}
\usepackage{booktabs}
\usepackage{enumitem}  
\usepackage{listings}  
\usepackage{xcolor}
\usepackage{tcolorbox}
\tcbuselibrary{skins,breakable}

\definecolor{promptbg}{RGB}{245, 245, 255}      
\definecolor{promptframe}{RGB}{100, 100, 150}   
\definecolor{outputbg}{RGB}{255, 252, 240}      
\definecolor{outputframe}{RGB}{150, 130, 80}    
\definecolor{codebg}{RGB}{248, 248, 248}        

\newtcolorbox{promptbox}[1]{
    breakable,
  colback=promptbg,
  colframe=promptframe,
  coltitle=white,
  colbacktitle=promptframe,
  title={\small\bfseries #1},
  fonttitle=\sffamily,
  arc=2mm,
  boxrule=0.5pt,
  left=4pt,
  right=4pt,
  top=4pt,
  bottom=4pt,
  enhanced,
  before=\par\vspace{6pt}\noindent,
  after=\par\vspace{6pt}
}

\newtcolorbox{outputbox}[1]{
breakable,
  colback=outputbg,
  colframe=outputframe,
  coltitle=white,
  colbacktitle=outputframe,
  title={\small\bfseries #1},
  fonttitle=\sffamily,
  arc=2mm,
  boxrule=0.5pt,
  left=4pt,
  right=4pt,
  top=4pt,
  bottom=4pt,
  enhanced,
  before=\par\vspace{6pt}\noindent,
  after=\par\vspace{6pt}
}

\definecolor{examplebg}{RGB}{240, 248, 255}      
\definecolor{exampleframe}{RGB}{70, 130, 180}    
\definecolor{successbg}{RGB}{240, 255, 240}      
\definecolor{successframe}{RGB}{60, 140, 60}     
\definecolor{partialbg}{RGB}{255, 250, 240}      
\definecolor{partialframe}{RGB}{200, 150, 50}    
\definecolor{failurebg}{RGB}{255, 245, 245}      
\definecolor{failureframe}{RGB}{180, 80, 80}     

\newtcolorbox{successbox}[1]{
breakable,
  colback=successbg,
  colframe=successframe,
  coltitle=white,
  colbacktitle=successframe,
  title={\small\bfseries #1},
  fonttitle=\sffamily,
  arc=2mm,
  boxrule=0.5pt,
  left=4pt, right=4pt, top=4pt, bottom=4pt,
  enhanced,
  before=\par\vspace{6pt}\noindent,
  after=\par\vspace{6pt}
}

\newtcolorbox{partialbox}[1]{
breakable,
  colback=partialbg,
  colframe=partialframe,
  coltitle=white,
  colbacktitle=partialframe,
  title={\small\bfseries #1},
  fonttitle=\sffamily,
  arc=2mm,
  boxrule=0.5pt,
  left=4pt, right=4pt, top=4pt, bottom=4pt,
  enhanced,
  before=\par\vspace{6pt}\noindent,
  after=\par\vspace{6pt}
}

\newtcolorbox{failurebox}[1]{
breakable,
  colback=failurebg,
  colframe=failureframe,
  coltitle=white,
  colbacktitle=failureframe,
  title={\small\bfseries #1},
  fonttitle=\sffamily,
  arc=2mm,
  boxrule=0.5pt,
  left=4pt, right=4pt, top=4pt, bottom=4pt,
  enhanced,
  before=\par\vspace{6pt}\noindent,
  after=\par\vspace{6pt}
}

\title{Schema-Free Text-to-SQL: \\ Learning Enterprise Database Structure from Query Logs}

\author{Ömer Faruk Akgül \thanks{Work done during an internship at Amazon.} \\
  University of Southern California \\\And
  Nikhil Kapahi \\
  Amazon \\\And
  Prajit Reddy Muppidi \\ 
  Amazon \\}

\begin{document}
\maketitle
\begin{abstract}
Text-to-SQL systems achieve remarkable accuracy on academic benchmarks, yet performance drops dramatically in enterprise deployments. This gap stems from a fundamental assumption: existing approaches require complete schema documentation, but enterprise databases typically lack comprehensive documentation, featuring cryptic naming and implicit relationships. We present the first text-to-SQL framework that operates without schema documentation, instead learning database structure from historical query logs. Query logs encode implicit structural knowledge: which tables are queried together, how they join, and what columns map to business concepts. We construct a knowledge graph from parsed query patterns and introduce entity-driven GraphRAG retrieval that grounds user questions in domain terminology rather than relying solely on embedding similarity. Validated on both an industrial deployment with expert evaluation and the Beaver benchmark, our approach substantially outperforms similarity-based retrieval methods, the dominant industry approach, in both executability and answer quality. On Beaver, we achieve 95.7\% execution validity compared to 60.8\% for embedding-based retrieval, with expert data engineers preferring our system 70\% to 10\% on internal deployment. Our work demonstrates that text-to-SQL can operate effectively in schema-ambiguous enterprise environments, enabling natural language database interfaces ``in the wild.''
\end{abstract}

\section{Introduction}

\begin{figure}[t]
\centering
\includegraphics[width=\columnwidth]{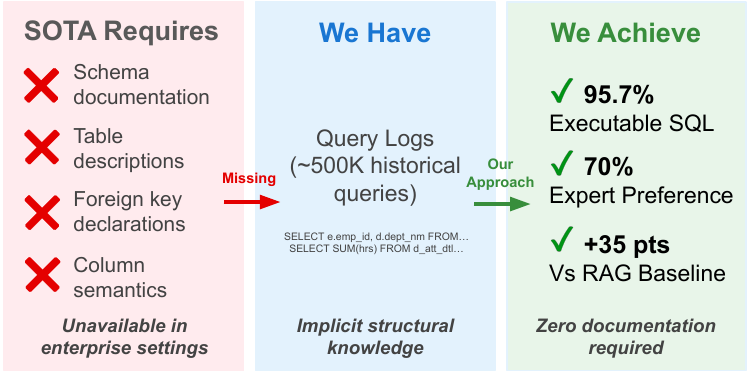}
\caption{State-of-the-art text-to-SQL systems require complete schema documentation unavailable in enterprise settings. Operating with only historical query logs, our approach achieves 95.7\% execution validity and 70\% expert preference over embedding-based retrieval.}
\label{fig:teaser}
\vspace{-2mm}
\end{figure}

Text-to-SQL systems have achieved remarkable performance on academic benchmarks, with state-of-the-art methods~\cite{shkapenyuk2025automaticmetadataextractiontexttosql, lee2024mcssqlleveragingmultipleprompts, gao2023texttosqlempoweredlargelanguage} achieving high accuracy on datasets like Spider~\cite{yu2018spider} and BIRD~\cite{li2023bird}. Yet these same systems struggle dramatically when deployed in enterprise environments~\cite{lei2025spider2,chen2025beaver}. This performance gap reflects a fundamental assumption mismatch: current text-to-SQL approaches universally require database identity known upfront, complete table schemas as input, explicit foreign key relationships, and interpretable column names. Enterprise databases offer none of these.

In practice, enterprise data ecosystems present what we term the \textit{schema ambiguity problem}. Tables and columns lack meaningful descriptions, with names employing domain-specific abbreviations that obscure their meaning to anyone outside the immediate team. Foreign keys are rarely declared explicitly; join relationships exist only in the institutional knowledge of database administrators or embedded in historical query patterns. These environments span hundreds to thousands of tables~\cite{lei2025spider2,chen2025beaver}, and schemas evolve continuously as business needs change, rendering any static documentation quickly outdated. These challenges are not edge cases but the norm in enterprise settings, and recent work on enterprise-oriented benchmarks has begun to quantify this gap~\cite{lei2025spider2,chen2025rubiksql}.

\paragraph{Our Insight.} We observe that while explicit documentation may be absent, enterprise databases contain an underutilized source of structural knowledge: historical query logs. Production SQL queries encode implicit information about database structure, revealing which tables are frequently queried together, how tables join in practice, and what columns map to specific business concepts. Query logs capture what documentation \textit{should} contain, learned from usage patterns rather than manual annotation.

Based on this insight, we present a text-to-SQL framework that operates without schema documentation, instead constructing a knowledge graph directly from query log analysis. State-of-the-art text-to-SQL systems are not designed for this setting, as they assume complete, human-readable schemas are available as input. Since state-of-the-art text-to-SQL systems are not designed for this setting, practitioners often turn to retrieval-augmented generation (RAG), which uses embedding-based similarity to narrow candidate tables and columns~\cite{zhang-etal-2023-refsql}. However, embedding similarity fundamentally struggles in enterprise settings: users ask about "approved time-off requests" and "hours worked" while schemas contain identifiers like \texttt{f\_pto\_agg} and \texttt{d\_att\_dtl}. This vocabulary mismatch cannot be bridged by semantic similarity alone~\cite{chen2025beaver}. To address this, we introduce entity-driven GraphRAG retrieval: we extract domain entities from user questions and traverse the knowledge graph to identify relevant tables, leveraging historical join patterns to ensure structural coherence.

Enterprise text-to-SQL also demands rethinking evaluation methodology. Open-ended analytical questions frequently admit multiple valid SQL formulations, rendering traditional exact-match metrics inadequate~\cite{lei2025spider2,chen2025beaver}. A query asking for ``sales trends'' might be correctly answered by grouping on weeks or months, selecting different but equally valid column subsets, or joining through alternative paths. We address this through multi-faceted evaluation combining execution-based accuracy~\cite{lei2025spider2}, blind pairwise comparison via LLM-as-judge validated by 83\% agreement with expert data engineers, and rubric-based analysis that decomposes SQL quality into interpretable dimensions. Our generated-versus-gold analysis (\S\ref{sec:gold_analysis}) further demonstrates that on open-ended questions, system-generated SQL achieves quality scores matching or exceeding human-written gold standards.

We make the following \textbf{contributions}:

\begin{enumerate}
    \item The first text-to-SQL framework that operates without schema documentation, learning database structure from historical query logs.
    
    \item Entity-driven GraphRAG retrieval that bridges the vocabulary gap between business terminology and cryptic schema identifiers through domain entity matching.
    
    \item A multi-faceted evaluation framework addressing open-ended question ambiguity through execution-based metrics, expert-validated LLM judging, and rubric-based analysis.
    
    \item Validation on an industrial deployment (240 tables, 70\% expert preference) and Beaver benchmark (95.7\% vs.\ 60.8\% execution validity), with ablations confirming entity matching as the key contributor.
\end{enumerate}

Our work demonstrates that text-to-SQL can operate effectively in schema-ambiguous enterprise environments, providing a new paradigm for natural language database interfaces ``in the wild.''
\section{Related Work}
Text-to-SQL systems achieve strong performance through in-context learning~\cite{gao2023texttosqlempoweredlargelanguage,dong2023c3} and fine-tuning~\cite{li2023resdsql,li2024codes}, with CHASE-SQL~\cite{pourreza2024chasesqlmultipathreasoningpreference} and MCS-SQL~\cite{lee2024mcssqlleveragingmultipleprompts} reaching near-perfect benchmark accuracy. However, these approaches assume complete schema documentation. Enterprise benchmarks including Spider~2.0~\cite{lei2025spider2} and BEAVER~\cite{chen2025beaver} reveal degradation under realistic conditions, while RubikSQL~\cite{chen2025rubiksql} highlights the need for domain knowledge in industrial settings.

Retrieval-augmented approaches address schema scale through embedding similarity~\cite{zhang-etal-2023-refsql,eben2025rasl}, but treat schema elements independently without capturing structural relationships. Graph-based methods model structure through relational encodings~\cite{wang2019rat,cao2021lgesql}, and Graph RAG~\cite{edge2024local} organizes knowledge into structured graphs, though none leverage query logs. Our approach builds on database workload analysis~\cite{chaudhuri1998autoadmin}, adapting log mining to extract semantic rather than physical structure.

\begin{figure*}[t]
\centering
\includegraphics[width=1.8\columnwidth]{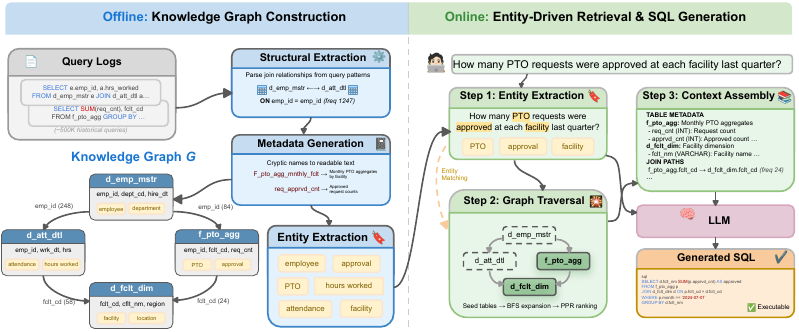}
\caption{System architecture. Offline: query logs are processed to construct a knowledge graph encoding table metadata, domain entities, and join relationships. Online: given a question, we extract entities, traverse the graph to retrieve relevant context, and generate SQL.}
\label{fig:framework}
\vspace{-4mm}
\end{figure*}

\section{Task Formulation}

Traditional text-to-SQL assumes access to a complete database schema $S = (T, C, FK)$ consisting of tables $T$, columns $C$, and foreign key relationships $FK$. Given a natural language question $Q$ and schema $S$, the task is to generate a SQL query $Y$ that answers the question:
\begin{equation}
Y = f(Q, S)
\end{equation}
This formulation presupposes that $S$ is fully documented and provided as input. Our setting differs: we have no schema documentation, only a corpus of historical query logs $L$.

We decompose the problem into three stages. First, we construct a knowledge graph $G$ from the query logs $L$ (offline, before any questions arrive). The graph encodes table metadata, domain entities associated with each table, and join relationships weighted by their frequency in historical queries. Second, given a question $Q$, we retrieve relevant context $\hat{C}$ from $G$ containing the tables, relationships, and metadata needed to answer the question. Third, we generate the SQL query:
\begin{equation}
Y = f(Q, \hat{C})
\end{equation}

This formulation subsumes the traditional setting: a documented schema $S$ is a degenerate case where structure is provided explicitly rather than inferred from usage patterns. The reverse does not hold. Query logs capture which tables are joined in practice, which columns co-occur, and how domain terminology maps to schema elements. This implicit knowledge cannot be recovered from schema documentation alone.

\section{Method}

We instantiate the three-stage framework introduced in Task Formulation: knowledge graph construction, entity-driven retrieval, and SQL generation. Figure~\ref{fig:framework} illustrates the complete pipeline.

\subsection{Knowledge Graph Construction}

Given a corpus of historical query logs $L = \{l_1, l_2, \ldots, l_n\}$, we construct a knowledge graph $G = (V, E)$ through three steps: structural extraction, metadata generation, and entity annotation.

\paragraph{Structural Extraction.}
We parse each query $l_i$ to identify table references and compute usage frequency:
\begin{equation}
\text{freq}(T) = \sum_{i=1}^{n} \mathbbm{1}[T \in l_i]
\end{equation}
where $\mathbbm{1}[\cdot]$ is the indicator function. We prioritize tables exceeding a frequency threshold, as these represent core entities in the database ecosystem.

Extracting join relationships from enterprise SQL requires more than syntactic parsing. Queries often contain implicit joins in WHERE clauses, relationships mediated through CTEs, and complex multi-way joins. We prompt an LLM to analyze each query and extract relationship triplets (Appendix~\ref{app:triplet_prompt}):
\begin{equation}
J(T_j, T_k) = \{(c_j, c_k, \tau)\}
\end{equation}
where $c_j, c_k$ are join columns and $\tau$ is the join type. The frequency of each join pattern serves as edge weight in $G$.

\paragraph{Metadata Generation.}
The logs reveal which tables and columns exist, but not what they mean. Names like \texttt{d\_wfm\_fclty\_pop\_dfntn\_mnl} offer little semantic guidance. We prompt an LLM with representative queries referencing each table $T$ to generate synthetic metadata $M(T)$ describing table purpose, column semantics, and typical usage patterns (Appendix~\ref{app:metadata_prompt}).

\paragraph{Entity Extraction.}
From the generated metadata $M(T)$, we extract domain entities $E(T)$ representing business concepts associated with each table (Appendix~\ref{app:entity_prompt}). For example, entities for an attendance table might include \{\textit{employee}, \textit{time-off}, \textit{approval rate}\}. These entities bridge the vocabulary gap between user questions and schema elements.

The resulting graph $G$ has nodes $V$ representing tables (annotated with metadata and entities) and edges $E$ representing join relationships (weighted by frequency). \

\subsection{Entity-Driven Retrieval}

Given a question $Q$, we retrieve relevant context $\hat{C}$ from $G$ through entity matching, graph traversal, and structural ranking.

\paragraph{Entity Matching.}
Embedding similarity between questions and table descriptions struggles with domain-specific terminology~\cite{chen2025beaver}. Users ask about ``shift swaps'' or ``approval rates,'' not \texttt{d\_shft\_swp\_agg}. We extract entities $E(Q)$ from the question using the same extraction process applied during graph construction, then match against table entities:
\begin{equation}
s_{\text{entity}}(T, Q) = \frac{|E(Q) \cap E(T)|}{|E(Q)|}
\end{equation}
Tables with high entity overlap become seed nodes for graph traversal.

\paragraph{Multi-Strategy Scoring.}
Entity matching alone may miss relevant tables when terminology varies. We combine three signals:
\begin{equation}
s(T, Q) = \alpha \cdot s_{\text{ent}} + \beta \cdot s_{\text{emb}} + \gamma \cdot s_{\text{hist}}
\end{equation}
where $s_{\text{ent}} = s_{\text{entity}}(T, Q)$, $s_{\text{emb}}$ measures embedding similarity between $Q$ and $M(T)$, and $s_{\text{hist}}$ scores tables appearing in historical queries similar to $Q$. We set $\alpha > \beta > \gamma$ to prioritize entity matching while retaining fallback strategies.

\paragraph{Graph Traversal.}
Starting from high-scoring seed tables, we traverse $G$ to discover additional tables connected through join relationships. We perform breadth-first search up to $k$ hops, scoring each path $\pi$ by entity coverage and join frequency:
\begin{equation}
\text{score}(\pi) = \sum_{T \in \pi} |E(Q) \cap E(T)| + \lambda \sum_{e \in \pi} w(e)
\end{equation}
where $w(e)$ is the edge weight (join frequency) and $\lambda$ balances entity coverage against established join patterns. This discovers tables not directly mentioned in the question but necessary for complete SQL construction.

\paragraph{Structural Ranking.}
To identify bridge tables connecting query-relevant entities, we apply Personalized PageRank with seed tables as the personalization vector. Tables with high PPR scores serve structural roles in the database even if not directly matched by entities.

\subsection{Context Assembly and SQL Generation}

The retrieval stage yields a set of relevant tables with their metadata, entity mappings, and join paths. We assemble these into a structured context $\hat{C}$ for SQL generation.

\paragraph{Context Structure.}
The context $\hat{C}$ comprises three components: (1) table metadata including descriptions and column information, (2) join paths derived from historical patterns, and (3) similar historical queries serving as few-shot examples (Appendix~\ref{app:context_format}). This hierarchical organization foregrounds the most relevant information for the LLM.

\paragraph{SQL Generation.}
We prompt an LLM with the question $Q$ and assembled context $\hat{C}$ to generate SQL (Appendix~\ref{app:sql_generation_prompt}):
\begin{equation}
Y = \text{LLM}(Q, \hat{C}, P_{\text{sql}})
\end{equation}
where $P_{\text{sql}}$ instructs the model to use only tables and columns present in the provided metadata, implement joins according to the identified paths, and avoid hallucinating schema elements. This grounding in retrieved context is critical: without it, LLMs frequently generate plausible but invalid SQL referencing non-existent columns~\cite{chen2025beaver}.

\subsection{Implementation}

We process query logs from an enterprise production system spanning 240 tables. Knowledge graph construction uses Claude 3.7 Sonnet (temperature 0) for metadata generation and entity extraction, implemented in NetworkX. Entity matching combines exact match ($\alpha=0.5$), substring match ($\beta=0.3$), and semantic similarity ($\gamma=0.2$) using E5-base-v2. Graph traversal is limited to $k=5$ hops.
\section{Evaluation Framework}
\label{sec:evaluation}

Traditional text-to-SQL metrics assume a single correct answer, which holds for deterministic questions with fully specified constraints. However, enterprise questions are frequently underspecified: Spider~2.0 notes that ``questions do not explicitly specify which columns to return''~\cite{lei2025spider2}, and BEAVER observes that real analyst questions admit multiple valid SQL formulations~\cite{chen2025beaver}. We therefore adopt a multi-faceted evaluation that extends traditional metrics to handle open-ended questions, combining execution-based metrics with LLM-as-judge assessment validated against human expert annotations.

\vspace{-2mm}

\paragraph{Metrics.}
We evaluate along three dimensions. \textbf{Execution validity} measures whether SQL runs without errors, providing a necessary but insufficient quality signal. \textbf{LLM-as-judge pairwise comparison} presents two SQL queries blindly (randomized A/B assignment) and scores executability, schema validity, filter correctness, join relevance, and result fidelity (Appendix~\ref{app:judge_prompt}). We validate this automated evaluation against annotations from two senior data engineers on our Amazon deployment, finding 83\% agreement on cases with clear human preference (\S\ref{sec:human_validation}). \textbf{Rubric-based scoring} decomposes SQL quality into interpretable dimensions for fine-grained analysis.

\vspace{-2mm}

\paragraph{Baseline.}
State-of-the-art text-to-SQL systems require complete schema information unavailable in our setting, and even with reconstructed documentation, their schema linking struggles without historical usage patterns (Appendix~\ref{app:dinsql}). We compare against an embedding-based RAG baseline operating under identical constraints, with implementation details in \S\ref{sec:experiments}.
\section{Experiments}
\label{sec:experiments}

We evaluate on two settings: an internal Amazon deployment with expert validation, and the public BEAVER benchmark for reproducibility.

\subsection{Experimental Setup}

We evaluate on two complementary settings: our Amazon deployment demonstrates real-world value with expert validation but cannot be independently reproduced, while BEAVER provides a public benchmark with gold SQL, enabling reproducible comparison and generated-vs-gold analysis.

\paragraph{Amazon Internal Deployment.}
We deploy on a production analytics database with 240 frequently-queried tables selected from thousands available, processing approximately 500K historical queries with zero schema documentation. The constructed knowledge graph contains 1,847 join edges extracted from query patterns. For evaluation, we use 35 test questions: 25 LLM-generated for coverage across table combinations, and 10 curated by subject matter experts reflecting real business needs. Two data engineers performed blind evaluation of SQL pairs.

\paragraph{BEAVER Benchmark.}
BEAVER~\cite{chen2025beaver} provides 209 questions across enterprise databases with 463 tables. Since BEAVER supplies schema files rather than query logs, we generate equivalent metadata from schema documentation to match our pipeline input format (Appendix~\ref{app:experimental_details}). Our adaptation discovers 1,113 implicit joins beyond the 127 declared relationships, yielding 1,310 total knowledge graph edges (Table~\ref{tab:beaver_kg_stats}).

\paragraph{Baseline.}
Our RAG baseline embeds table metadata using E5-base-v2, retrieves top-12 tables via cosine similarity, and generates SQL with Claude 3.7 Sonnet. Both systems receive identical generated metadata; the difference is retrieval strategy.

\subsection{Main Results}

Table~\ref{tab:main_results} summarizes performance on BEAVER. Our approach achieves 95.7\% execution validity compared to 60.8\% for RAG, a 35-point improvement demonstrating that knowledge graph structure produces coherent table-column combinations.

\begin{table}[t]
\centering
\small
\begin{tabular}{lcc}
\toprule
\textbf{Metric} & \textbf{Ours} & \textbf{RAG} \\
\midrule
Execution Validity & 200/209 (95.7\%) & 127/209 (60.8\%) \\
\midrule
LLM Judge Wins & 95 (45.5\%) & 71 (34.0\%) \\
Rubric Judge Wins & 100 (47.8\%) & 64 (30.6\%) \\
\bottomrule
\end{tabular}
\caption{BEAVER benchmark results (209 questions). Ties: 21 both adequate (LLM), 45 both (Rubric).}
\label{tab:main_results}
\vspace{-3mm}
\end{table}

In blind pairwise comparison, LLM judges prefer our system in 45.5\% of cases versus 34.0\% for RAG. The rubric-based judge shows consistent preference (47.8\% vs 30.6\%). Table~\ref{tab:dimensions} breaks down by dimension: our largest gains appear in executability (+74 vs +1) and schema validity (+60 vs +17), directly validating that KG-guided retrieval reduces hallucinated columns.

\begin{table}[t]
\centering
\small
\begin{tabular}{lccc}
\toprule
\textbf{Dimension} & \textbf{Ours} & \textbf{RAG} & \textbf{Both} \\
\midrule
Executability & 74 & 1 & 126 \\
Schema Validity & 60 & 17 & 132 \\
Filter Correctness & 52 & 28 & 120 \\
Join Relevance & 44 & 55 & 62 \\
Result Fidelity & 85 & 69 & 38 \\
\bottomrule
\end{tabular}
\caption{Rubric dimension wins on BEAVER test set.}
\label{tab:dimensions}
\vspace{-1mm}
\end{table}

\subsection{Human Validation}
\label{sec:human_validation}

To establish that LLM-as-judge reflects real-world utility, we compare judgments against expert annotations on our Amazon deployment. Table~\ref{tab:amazon} shows both data engineers and LLM judges strongly prefer our system (70\% vs 10\% for experts, 80\% vs 16\% for LLM). On questions where humans expressed clear preference, LLM judgments agreed in 83\% of cases, validating automated evaluation as a reliable proxy for expert assessment.

\begin{table}[t]
\centering
\small
\begin{tabular}{lcc}
\toprule
\textbf{Evaluator} & \textbf{Ours} & \textbf{RAG} \\
\midrule
Expert Consensus & 70\% & 10\% \\
LLM Judge & 80\% & 16\% \\
\midrule
LLM-Human Agreement & \multicolumn{2}{c}{83\%} \\
\bottomrule
\end{tabular}
\caption{Amazon evaluation. Remaining cases are ties.}
\label{tab:amazon}
\vspace{-4mm}
\end{table}

\subsection{Rethinking Gold Standards}
\label{sec:gold_analysis}

A key motivation for our evaluation framework is that enterprise questions often admit multiple valid SQL formulations, making ``gold'' answers one reasonable solution among many. To test this empirically, we compare our generated SQL against human-written gold SQL from BEAVER using the same blind evaluation protocol.

Table~\ref{tab:gold} reveals a striking pattern. On open-ended questions where interpretation varies, our system achieves quality scores that exceed gold (9.00 vs 8.85). In pairwise comparison, judges find our SQL adequate or better in 46\% of open-ended cases. On deterministic questions with clear correct answers, gold appropriately dominates.

\begin{table}[t]
\centering
\small
\begin{tabular}{lcc}
\toprule
& \textbf{Deterministic} & \textbf{Open-Ended} \\
& (n=114) & (n=95) \\
\midrule
Gold Avg Score & 8.50 & 8.85 \\
Ours Avg Score & 7.55 & \textbf{9.00} \\
\midrule
Gold Wins & 64.9\% & 61.1\% \\
Ours Wins & 26.3\% & 28.4\% \\
Both Adequate & 3.5\% & 7.4\% \\
\bottomrule
\end{tabular}
\caption{Generated vs Gold SQL by question type. On open-ended questions, generated SQL matches or exceeds gold quality.}
\label{tab:gold}
\vspace{-1mm}
\end{table}

This finding validates our evaluation approach: for the substantial portion of enterprise questions that are inherently open-ended, traditional gold-based metrics penalize valid alternative formulations. SQL that judges rate 9.0/10 should not be considered wrong simply because it differs from one particular gold answer. Full breakdown by question type and database appears in Appendix~\ref{app:question_analysis}.

\subsection{Ablation Studies}

We isolate component contributions by removing each from the full pipeline (Table~\ref{tab:ablation}). Entity matching shows the largest impact: removing it drops scores by 5.4\% and loses 43\% of pairwise comparisons. This confirms that domain terminology requires explicit entity grounding beyond embedding similarity.

\begin{table}[t]
\centering
\small
\begin{tabular}{lccc}
\toprule
\textbf{Configuration} & \textbf{Score} & \textbf{$\Delta$} & \textbf{vs Full} \\
\midrule
Full Pipeline & 7.76 & -- & -- \\
$-$ Entity Matching & 7.34 & $-$5.4\% & 43\% / 21\% \\
$-$ KG Traversal & 7.60 & $-$2.1\% & 41\% / 28\% \\
$-$ PPR Ranking & 7.62 & $-$1.7\% & 21\% / 18\% \\
\bottomrule
\end{tabular}
\caption{Ablation on BEAVER. Score: open-ended evaluation (0--10). vs Full: pairwise wins (Full / Ablated).}
\label{tab:ablation}
\vspace{-4mm}
\end{table}

KG traversal contributes 2.1\%, validating that graph navigation discovers necessary tables not directly matched by entities. PPR ranking provides modest gains with 60.8\% ties, suggesting it refines rather than fundamentally changes retrieval. Per-dimension analysis (Appendix~\ref{app:ablation_details}) shows entity matching mostly impacts result fidelity, while KG traversal primarily affects join relevance.
\section{Conclusion}
\label{sec:conclusion}

We presented a text-to-SQL framework that operates without schema documentation by mining query logs to construct a knowledge graph of table relationships and domain entities. Entity-driven GraphRAG retrieval bridges the vocabulary gap between business terminology and cryptic schema identifiers, substantially outperforming embedding-based baselines. Our results demonstrate effective text-to-SQL in schema-ambiguous enterprise environments, while highlighting the need for evaluation beyond single-gold matching.
\section{Limitations}
\label{sec:limitation}

The BEAVER adaptation uses schema-derived rather than log-derived edges, which lack frequency information and may include relationships absent from actual usage patterns.


\bibliography{custom}

\clearpage

\appendix

\section{Appendix}
\label{sec:appendix}

\section{Prompts}
\label{app:prompts}

This appendix provides the prompts used in knowledge graph construction, SQL generation, and evaluation.

\subsection{Metadata Generation Prompt}
\label{app:metadata_prompt}

\begin{promptbox}{Metadata Generation}
Given the table \texttt{[TABLE\_NAME]} which appears in \texttt{[N]} queries, and the following representative queries:

\texttt{[SAMPLE\_QUERIES]}

Generate metadata describing:
\begin{enumerate}[leftmargin=*, itemsep=0pt, topsep=2pt]
\item Table purpose and business context
\item Column names, meanings, and likely data types
\item Primary/foreign key columns
\item Common join patterns with other tables
\item Typical filtering and aggregation patterns
\end{enumerate}

Format as JSON with fields: \texttt{table\_name}, \texttt{description}, \texttt{columns} (list with name, description, type), \texttt{key\_columns}, \texttt{common\_joins}, \texttt{common\_filters}.
\end{promptbox}

\subsection{Join Relationship Extraction Prompt}
\label{app:triplet_prompt}

\begin{promptbox}{Join Relationship Extraction}
Extract table join relationships from the following SQL query. Identify ALL relationships, including implicit ones.

\textbf{SQL Query:} \texttt{[SQL\_QUERY]}

\textbf{Tables referenced:} \texttt{[TABLE\_LIST]}

Consider the following relationship types:
\begin{enumerate}[leftmargin=*, itemsep=0pt, topsep=2pt]
\item Explicit joins (JOIN, LEFT JOIN, INNER JOIN)
\item Implicit joins in WHERE clauses
\item Relationships through CTEs where tables share keys
\item Logical data flow beyond syntax
\end{enumerate}

For each relationship, return: \texttt{left\_table}, \texttt{right\_table}, \texttt{join\_type}, \texttt{join\_conditions}, \texttt{relationship\_type} (EXPLICIT or IMPLICIT).
\end{promptbox}

\subsection{Entity Extraction Prompt}
\label{app:entity_prompt}

\begin{promptbox}{Entity Extraction}
You are an expert entity extractor for enterprise data systems.

\textbf{WHAT TO EXTRACT}
\begin{itemize}[leftmargin=*, itemsep=0pt, topsep=2pt]
\item Core business concepts (e.g., shift swap, approval rate, employee)
\item Processes (e.g., registration, scheduling, approvals)
\item Organizational entities (e.g., facility, department, site)
\item Primary metrics (e.g., counts, rates, hours)
\end{itemize}

\textbf{WHAT TO AVOID}
\begin{itemize}[leftmargin=*, itemsep=0pt, topsep=2pt]
\item Specific IDs or literal values
\item Dates or absolute time ranges
\item Generic database terms (table, record, row)
\end{itemize}

Extract concise noun phrases (1--3 words). Preserve business acronyms.

\textbf{Input:} \texttt{[TABLE\_METADATA]}

\textbf{Output:} Return only a valid JSON array of entity strings.
\end{promptbox}

\subsection{SQL Generation Prompt}
\label{app:sql_generation_prompt}

\begin{promptbox}{SQL Generation}
Generate a SQL query for the following question using ONLY the tables and columns in the provided context.

\textbf{Context:}
\texttt{[TABLE\_METADATA]}
\texttt{[JOIN\_PATHS]}
\texttt{[SIMILAR\_QUERIES]}

\textbf{Question:} \texttt{[USER\_QUESTION]}

\textbf{Instructions:}
\begin{enumerate}[leftmargin=*, itemsep=0pt, topsep=2pt]
\item Use ONLY tables and columns explicitly listed in the metadata
\item If a needed column is not in the metadata, explain the limitation
\item Follow join patterns from the provided join paths
\item Use table aliases consistently
\item Do NOT hallucinate tables or columns
\end{enumerate}
\end{promptbox}

\subsection{LLM-as-Judge Pairwise Comparison}
\label{app:judge_prompt}

We employ blind pairwise comparison where an LLM judge evaluates two SQL queries without knowing their source. A/B assignment is randomized across comparisons to prevent systematic bias.

\begin{promptbox}{Pairwise SQL Comparison}
Compare two SQL queries (blind evaluation). You are NOT told which system produced each SQL.

\textbf{Inputs:} \texttt{[QUESTION]}, \texttt{[SQL\_A]}, \texttt{[SQL\_B]}, \texttt{[EXEC\_A]}, \texttt{[EXEC\_B]}, \texttt{[METADATA]}

\textbf{Score 0--2 on four dimensions:}
\begin{enumerate}[leftmargin=*, itemsep=0pt, topsep=2pt]
\item \textbf{Schema Validity}: Tables/columns exist and are appropriate
\item \textbf{Filter Correctness}: WHERE/GROUP BY/LIMIT reflect constraints
\item \textbf{Join Relevance}: Joins are plausible and necessary
\item \textbf{Result Fidelity}: Output satisfies the information need
\end{enumerate}

If SQL fails to execute, score 0 on all dimensions.

\textbf{Return JSON}: \texttt{dimensions}, \texttt{total\_score\_A}, \texttt{total\_score\_B}, \texttt{overall\_winner} (A|B|both|neither), \texttt{overall\_rationale}.
\end{promptbox}

\subsection{Open-Ended Scoring Rubric}
\label{app:openended_prompt}

For questions where multiple interpretations are valid, we assess how well SQL addresses the question's intent rather than matching a specific gold formulation.

\begin{promptbox}{Open-Ended SQL Evaluation}
Score the SQL on a 0--10 scale. Focus on whether the SQL \textbf{fulfills the intent} of the question, not whether it exactly matches any gold answer.

\textbf{Question:} \texttt{[QUESTION]}

\textbf{SQL:} \texttt{[SQL]}

\textbf{Execution Result:} \texttt{[EXEC\_RESULT]}

\medskip
\textbf{Scoring rubric}:

\begin{enumerate}[leftmargin=*, itemsep=2pt, topsep=4pt]
\item \textbf{Executability} (0--2): 2=executes successfully, 1=minor fixable error, 0=fails
\item \textbf{Table/Join Relevance} (0--2): 2=clearly aligned, 1=partial, 0=unrelated
\item \textbf{Field Coverage} (0--3): 3=all key fields, 2=most, 1=some, 0=none
\item \textbf{Output Usefulness} (0--2): 2=answers question, 1=partial, 0=not useful
\item \textbf{Interpretation Quality} (0--1): 1=reasonable interpretation of ambiguous terms
\end{enumerate}

\textbf{Important}: For open-ended questions, reward reasonable interpretations.

\textbf{Return JSON}: \texttt{score}, \texttt{subscores}, \texttt{verdict} (yes|partial|no), \texttt{rationale}, \texttt{issues}.
\end{promptbox}

\subsection{Question Type Classification}
\label{app:question_classification}

We classify questions into two categories to enable stratified analysis of system performance under different levels of ambiguity:

\begin{promptbox}{Question Type Classification}
Classify \texttt{[QUESTION]} into one of two categories:

\begin{enumerate}[leftmargin=*, itemsep=2pt, topsep=4pt]
\item \textbf{Deterministic}: Fully specifies all constraints and required fields; essentially one correct SQL formulation exists.

\textit{Examples}: ``Total revenue for product ID 123 in Q3 2024?'' / ``List employees in department 5 hired before 2020.''

\item \textbf{Open-Ended}: Underspecified or ambiguous; multiple valid SQL formulations could satisfy the intent.

\textit{Examples}: ``Show top performing products.'' (How many? By what metric?) / ``Analyze attendance trends.'' (What granularity?)
\end{enumerate}

\textbf{Return}: Category name and brief justification.
\end{promptbox}

On BEAVER, this classification yields 114 deterministic questions and 95 open-ended questions. The prevalence of open-ended questions (45\%) motivates our multi-faceted evaluation framework, nearly half of enterprise questions cannot be fairly evaluated against a single gold standard.

\section{Data Formats}
\label{app:formats}

\subsection{Knowledge Graph Triplet Format}
\label{app:triplet_format}

Each edge in the knowledge graph stores join relationships as triplets:

\begin{outputbox}{Triplet Structure}
\begin{lstlisting}[basicstyle=\ttfamily\small]
{
  "left_table": "schema.employees",
  "right_table": "schema.attendance",
  "join_type": "INNER",
  "join_conditions": "employees.id = attendance.emp_id",
  "relationship_type": "EXPLICIT",
  "frequency": 342,
  "left_table_entities": ["employee", "worker"],
  "right_table_entities": ["attendance", "time-off"],
  "shared_entities": ["employee"]
}
\end{lstlisting}
\end{outputbox}

\subsection{Context Assembly Format}
\label{app:context_format}

The context $\hat{C}$ provided to the LLM for SQL generation is structured hierarchically:

\begin{outputbox}{Context Structure}
\textbf{1. Table Metadata}
\begin{lstlisting}[basicstyle=\ttfamily\small]
Table: shift_swap_agg
Description: Aggregated shift swap metrics 
  by month and facility
Columns:
  - date_trunc_month: DATE | Month of record
  - facility_id: VARCHAR | Facility identifier
  - nbr_swap_request: INT | Total swap requests
  - nbr_swap_approved: INT | Approved requests
\end{lstlisting}

\textbf{2. Join Paths}
\begin{lstlisting}[basicstyle=\ttfamily\small]
shift_swap_agg.facility_id -> 
  facility_dim.facility_id (freq: 847)
\end{lstlisting}

\textbf{3. Similar Queries}
\begin{lstlisting}[basicstyle=\ttfamily\small]
Q: "What was the approval rate last month?"
SQL: SELECT SUM(nbr_swap_approved) * 100.0 / 
     SUM(nbr_swap_request) FROM shift_swap_agg
     WHERE date_trunc_month = DATE '2025-06-01'
\end{lstlisting}
\end{outputbox}

\section{BEAVER Benchmark Adaptation}
\label{app:beaver_adaptation}

BEAVER~\cite{chen2025beaver} provides schema files rather than query logs, requiring adaptation to evaluate our approach on a reproducible benchmark. This section details the conversion pipeline that enables controlled experiments while maintaining methodological consistency.

\subsection{Motivation for Adaptation}

Our primary evaluation on Amazon internal data demonstrates practical value but cannot be independently reproduced. BEAVER offers complementary benefits: (1) a public test set enabling reproducible comparison, (2) gold SQL annotations for measuring accuracy against expert solutions, and (3) diverse database domains (academic data warehouse, OpenStack infrastructure). However, BEAVER lacks the query logs central to our approach.

We address this gap by generating equivalent metadata from BEAVER's schema files, constructing a knowledge graph that mirrors the structure our pipeline would produce from logs. This tests whether entity-driven GraphRAG generalizes beyond the log-based setting while enabling direct comparison against gold SQL.

\subsection{Schema-to-Metadata Conversion}

BEAVER provides structured schema files containing table names, column definitions, data types, and declared foreign keys. We convert these to our metadata format through three steps:

\paragraph{Step 1: Table Metadata Generation.}
For each of BEAVER's 463 tables, we generate synthetic metadata using the same LLM prompting approach as our main pipeline. The prompt receives the table name, column names and types, and any schema descriptions provided by BEAVER. The output includes table purpose descriptions, column semantics, and inferred usage patterns.

\paragraph{Step 2: Join Relationship Discovery.}
BEAVER declares only 127 explicit foreign key relationships across its databases---far fewer than typical enterprise usage would reveal. To construct a realistic knowledge graph, we discover implicit joins through:
\begin{enumerate}[leftmargin=*, itemsep=1pt]
    \item \textbf{Column name matching}: Identify candidate pairs where columns share naming patterns (e.g., \texttt{*\_KEY}, \texttt{*\_ID})
    \item \textbf{LLM validation}: For each candidate pair, prompt an LLM to assess whether the relationship represents a valid join based on table semantics
    \item \textbf{Confidence filtering}: Accept joins with low/medium/high confidence ratings
\end{enumerate}

This process discovered 1,113 implicit joins beyond the 127 declared relationships, yielding a knowledge graph with 1,310 total edges across 463 tables. Table~\ref{tab:beaver_kg_stats} summarizes the construction statistics.

\begin{table}[h]
\centering
\small
\begin{tabular}{lr}
\toprule
\textbf{Statistic} & \textbf{Value} \\
\midrule
Total tables & 463 \\
Explicit FK relationships & 127 \\
Implicit joins discovered & 1,113 \\
Total KG edges & 1,310 \\
Unique entities extracted & 1,844 \\
\bottomrule
\end{tabular}
\caption{BEAVER knowledge graph construction statistics.}
\label{tab:beaver_kg_stats}
\end{table}

\paragraph{Step 3: Entity Extraction.}
We extract domain entities from generated metadata using the same prompt as our main pipeline (Appendix~\ref{app:entity_prompt}). Entities are normalized to lowercase and deduplicated. The resulting 1,844 unique entities enable the entity-driven retrieval that differentiates our approach from similarity-based baselines.

\subsection{Implicit Join Validation}

The implicit join discovery uses an LLM to validate candidate relationships:

\begin{promptbox}{Implicit Join Validation}
Assess whether two tables should be joined based on shared column names and table semantics.

\textbf{Left Table:} \texttt{[LEFT\_TABLE\_NAME]}

\texttt{[LEFT\_TABLE\_METADATA]}

\textbf{Right Table:} \texttt{[RIGHT\_TABLE\_NAME]}

\texttt{[RIGHT\_TABLE\_METADATA]}

\textbf{Candidate Join Column:} \texttt{[COLUMN\_NAME]}

Determine if this represents a valid join relationship. Consider:
\begin{itemize}[leftmargin=*, itemsep=0pt]
\item Do the tables share a logical relationship?
\item Does the column semantically connect both tables?
\item Would this join be meaningful for analytical queries?
\end{itemize}

\textbf{Return JSON}: \texttt{is\_join} (boolean), \texttt{confidence} (low|medium|high), \texttt{explanation}.
\end{promptbox}

\subsection{Limitations of Adaptation}

The schema-based adaptation differs from log-based construction in important ways:
\begin{itemize}[leftmargin=*, itemsep=1pt]
    \item \textbf{No frequency information}: All edges receive equal weight (1.0) rather than usage-derived frequencies
    \item \textbf{Synthetic relationships}: LLM-validated joins may include relationships that would never appear in actual queries
    \item \textbf{Missing query patterns}: No historical queries available for few-shot examples
\end{itemize}

Despite these limitations, BEAVER results demonstrate that entity-driven GraphRAG provides value even without query logs, suggesting the approach generalizes beyond the log-based setting.

\section{Experimental Details}
\label{app:experimental_details}

\subsection{Dataset Statistics}

Table~\ref{tab:dataset_stats} summarizes both evaluation settings.

\begin{table}[h]
\centering
\small
\begin{tabular}{lcc}
\toprule
& \textbf{Amazon} & \textbf{BEAVER} \\
\midrule
Total Tables & 240 & 463 \\
Query Logs & $\sim$500K & -- \\
KG Join Edges & 1,847 & 1,310 \\
Test Questions & 35 & 209 \\
\quad Deterministic & -- & 114 \\
\quad Open-Ended & -- & 95 \\
Gold SQL Available & No & Yes \\
\bottomrule
\end{tabular}
\caption{Dataset statistics for both evaluation settings.}
\label{tab:dataset_stats}
\end{table}

\subsection{Table Selection Statistics}

Table~\ref{tab:table_selection} compares retrieval behavior between our system and the RAG baseline on BEAVER.

\begin{table}[h]
\centering
\small
\begin{tabular}{lcc}
\toprule
\textbf{Metric} & \textbf{Ours} & \textbf{RAG} \\
\midrule
Avg tables selected & 12.61 & 12.00 \\
Median & 13 & 12 \\
Min / Max & 5 / 16 & 12 / 12 \\
\% of total (463) & 2.72\% & 2.59\% \\
\bottomrule
\end{tabular}
\caption{Table selection statistics on BEAVER. Our system adaptively selects 5--16 tables based on query complexity; RAG uses fixed top-$k$=12.}
\label{tab:table_selection}
\end{table}

Our system's adaptive selection (5--16 tables) reflects query complexity, while RAG's fixed retrieval cannot adjust to simpler or more complex questions.

\subsection{Ablation Configuration Details}
\label{app:ablation_details}

Each ablation removes a single component while keeping others intact:
\begin{itemize}[leftmargin=*, itemsep=1pt]
\item \textbf{$-$ Entity Matching}: Relies only on embedding similarity; entity overlap score set to zero.
\item \textbf{$-$ KG Traversal}: Retrieves only directly-matched tables; disables BFS from seed nodes.
\item \textbf{$-$ PPR Ranking}: Uses raw scores without Personalized PageRank reranking.
\end{itemize}

Table~\ref{tab:ablation_dims} shows per-dimension rubric analysis for ablations.

\begin{table}[h]
\centering
\small
\begin{tabular}{lccccc}
\toprule
\textbf{Ablation} & \textbf{Exec} & \textbf{Schema} & \textbf{Filter} & \textbf{Join} & \textbf{Result} \\
\midrule
$-$ Entity & 16/11 & 21/11 & 28/19 & 37/22 & 75/42 \\
$-$ KG Trav & 16/10 & 28/13 & 33/21 & 51/36 & 78/54 \\
$-$ PPR & 13/12 & 14/10 & 15/14 & 17/12 & 41/34 \\
\bottomrule
\end{tabular}
\caption{Ablation dimension breakdown (Full wins / Ablated wins). Entity matching has largest impact on result fidelity.}
\label{tab:ablation_dims}
\end{table}

\section{Question Type Analysis}
\label{app:question_analysis}

\subsection{Generated vs Gold by Question Type}

Table~\ref{tab:gold_detail} shows detailed comparison by question type. On deterministic questions, gold SQL appropriately dominates. On open-ended questions, our system achieves comparable or superior quality, supporting our argument that traditional gold-based evaluation inadequately captures valid alternative formulations.

\begin{table}[h]
\centering
\small
\begin{tabular}{lcccc}
\toprule
& \multicolumn{2}{c}{\textbf{Deterministic}} & \multicolumn{2}{c}{\textbf{Open-Ended}} \\
& Gold & Ours & Gold & Ours \\
\midrule
Avg Score & 8.50 & 7.55 & 8.85 & \textbf{9.00} \\
\midrule
Blind Judge Wins & 74 & 30 & 58 & 27 \\
Both Adequate & \multicolumn{2}{c}{4} & \multicolumn{2}{c}{7} \\
\midrule
Rubric Judge Wins & 65 & 39 & 51 & 33 \\
Both Adequate & \multicolumn{2}{c}{10} & \multicolumn{2}{c}{11} \\
\bottomrule
\end{tabular}
\caption{Generated vs Gold comparison by question type.}
\label{tab:gold_detail}
\end{table}

\subsection{Ours vs RAG by Question Type}

Table~\ref{tab:rag_detail} shows consistent improvements across both question types, with larger margins on open-ended questions where entity grounding provides greater benefit.

\begin{table}[h]
\centering
\small
\begin{tabular}{lcccc}
\toprule
& \multicolumn{2}{c}{\textbf{Deterministic}} & \multicolumn{2}{c}{\textbf{Open-Ended}} \\
& Ours & RAG & Ours & RAG \\
\midrule
Avg Score & 7.54 & 6.76 & 8.99 & 8.65 \\
Exec Valid & 95.6\% & 58.8\% & 96.8\% & 63.2\% \\
\bottomrule
\end{tabular}
\caption{Ours vs RAG by question type.}
\label{tab:rag_detail}
\end{table}

\subsection{Per-Database Results}

Table~\ref{tab:per_db} shows consistent improvements across BEAVER's databases, with largest gains on the complex data warehouse (dw).

\begin{table}[h]
\centering
\small
\begin{tabular}{lcccc}
\toprule
\textbf{Database} & \textbf{n} & \textbf{Ours Exec} & \textbf{RAG Exec} & \textbf{$\Delta$} \\
\midrule
dw & 121 & 94.2\% & 47.1\% & +47.1 \\
csail\_nova & 43 & 100\% & 72.1\% & +27.9 \\
csail\_neutron & 30 & 100\% & 93.3\% & +6.7 \\
keystone & 8 & 100\% & 100\% & 0 \\
Other & 7 & 100\% & 100\% & 0 \\
\bottomrule
\end{tabular}
\caption{Execution validity by database.}
\label{tab:per_db}
\end{table}

\section{Example Queries with Rubric Analysis}
\label{app:examples}

We present representative examples from Beaver generations, spanning high, medium, and low quality generations with full LLM rubric evaluations.

\subsection{High Quality: Entity Matching Enables Correct Schema Selection}

\begin{successbox}{Example 1: Material Status Analysis --- Score: 10/10}
\textbf{Question:} What is material status, the total number of materials, subjects, schools, and most recent publication year for each material status?

\textbf{Question Type:} Deterministic

\textbf{Generated SQL:}
\begin{lstlisting}[basicstyle=\ttfamily\scriptsize]
SELECT tms.TIP_MATERIAL_STATUS,
  COUNT(DISTINCT tm.TIP_MATERIAL_KEY) AS TotalMaterials,
  COUNT(DISTINCT td.SUBJECT_ID) AS TotalSubjects,
  COUNT(DISTINCT tso.OFFER_SCHOOL_NAME) AS TotalSchools,
  MAX(tm.YEAR) AS MostRecentPublicationYear
FROM dw.TIP_MATERIAL_STATUS tms
LEFT JOIN dw.TIP_DETAIL td 
  ON tms.TIP_MATERIAL_STATUS_KEY = td.TIP_MATERIAL_STATUS_KEY
LEFT JOIN dw.TIP_MATERIAL tm 
  ON td.TIP_MATERIAL_KEY = tm.TIP_MATERIAL_KEY
LEFT JOIN dw.TIP_SUBJECT_OFFERED tso 
  ON td.TIP_SUBJECT_OFFERED_KEY = tso.TIP_SUBJECT_OFFERED_KEY
GROUP BY tms.TIP_MATERIAL_STATUS
\end{lstlisting}

{\footnotesize
\begin{tabular}{@{}llp{3.4cm}@{}}
\toprule
\textbf{Dimension} & \textbf{Score} & \textbf{Assessment} \\
\midrule
Executability & 2/2 & Executes successfully \\
Schema Validity & 2/2 & All tables/columns correct \\
Filter Correctness & 2/2 & GROUP BY properly applied \\
Join Relevance & 2/2 & Four-table join path correct \\
Result Fidelity & 2/2 & All requested fields returned \\
\midrule
\textbf{Total} & \textbf{10/10} & \\
\bottomrule
\end{tabular}
}

\textbf{Verdict:} \textit{yes} --- fully answers the question.
\end{successbox}

\paragraph{Analysis.} Entity matching on ``material status'' $\rightarrow$ \texttt{TIP\_MATERIAL\_STATUS}, ``subjects'' $\rightarrow$ \texttt{TIP\_SUBJECT\_OFFERED}, and ``schools'' $\rightarrow$ \texttt{OFFER\_SCHOOL\_NAME} guided retrieval to correct schema elements. The knowledge graph provided the join path through \texttt{TIP\_DETAIL} as a bridge table.

\subsection{High Quality: KG Traversal Discovers Non-Obvious Tables}

\begin{successbox}{Example 2: Mailing List Visibility --- Score: 10/10}
\textbf{Question:} For each mailing list, list its owner, owner type, member visibility (`Public Members' if public, `Hidden Members' otherwise), and member count per visibility type. Include grand totals per owner.

\textbf{Question Type:} Deterministic

\textbf{Generated SQL:}
\begin{lstlisting}[basicstyle=\ttfamily\scriptsize]
WITH ListVisibility AS (
  SELECT ml.MOIRA_LIST_NAME, mld.MOIRA_LIST_KEY,
    mlo.OWNER, mlo.OWNER_TYPE,
    CASE WHEN ml.IS_PUBLIC = 'Y' THEN 'Public Members'
         ELSE 'Hidden Members' END AS MemberVisibility
  FROM dw.MOIRA_LIST AS ml
  JOIN dw.MOIRA_LIST_DETAIL AS mld 
    ON ml.MOIRA_LIST_KEY = mld.MOIRA_LIST_KEY
  JOIN dw.MOIRA_LIST_OWNER AS mlo 
    ON mld.MOIRA_LIST_OWNER_KEY = mlo.MOIRA_LIST_OWNER_KEY
  WHERE ml.IS_MOIRA_MAILING_LIST = 'Y'
), ListVisibilityCounts AS (
  SELECT OWNER, OWNER_TYPE, MemberVisibility, 
         COUNT(*) AS MemberCount
  FROM ListVisibility
  GROUP BY OWNER, OWNER_TYPE, MemberVisibility
), GrandTotal AS (
  SELECT OWNER, OWNER_TYPE, NULL AS MemberVisibility,
         SUM(MemberCount) AS MemberCount
  FROM ListVisibilityCounts GROUP BY OWNER, OWNER_TYPE
)
SELECT * FROM ListVisibilityCounts
UNION ALL SELECT * FROM GrandTotal
ORDER BY OWNER, OWNER_TYPE, MemberVisibility;
\end{lstlisting}

{\footnotesize
\begin{tabular}{@{}llp{3.4cm}@{}}
\toprule
\textbf{Dimension} & \textbf{Score} & \textbf{Assessment} \\
\midrule
Executability & 2/2 & Complex CTEs execute correctly \\
Schema Validity & 2/2 & MOIRA tables correctly identified \\
Filter Correctness & 2/2 & IS\_MOIRA\_MAILING \_LIST filter \\
Join Relevance & 2/2 & Three-table join path appropriate \\
Result Fidelity & 2/2 & Grand totals included as requested \\
\midrule
\textbf{Total} & \textbf{10/10} & \\
\bottomrule
\end{tabular}
}

\textbf{Verdict:} \textit{yes} --- comprehensive answer with correct structure.
\end{successbox}

\paragraph{Analysis.} The question mentions ``mailing list,'' ``owner,'' and ``visibility'' but not tables \texttt{MOIRA\_LIST\_DETAIL} or \texttt{MOIRA\_LIST\_OWNER}. Entity extraction identified ``mailing list'' $\rightarrow$ \texttt{MOIRA\_LIST}, and KG traversal discovered related tables through join edges.

\subsection{Medium Quality: Reasonable Interpretation of Ambiguous Question}

\begin{partialbox}{Example 3: Computer Science Room Query --- Score: 8/10}
\textbf{Question:} List unique full room names, building names, addresses, cities, states, postal codes, and building heights for rooms associated with subjects Computer Science students can enroll in.

\textbf{Question Type:} Open-ended (``CS students can enroll'' is ambiguous)

\textbf{Generated SQL:}
\begin{lstlisting}[basicstyle=\ttfamily\scriptsize]
SELECT DISTINCT fr.ROOM_FULL_NAME, b.BUILDING_NAME,
  fba.STREET_NUMBER || ' ' || fba.STREET_NAME || ' ' || 
    fba.STREET_SUFFIX AS STREET_ADDRESS,
  fba.CITY, fba.STATE, fba.POSTAL_CODE, fb.BUILDING_HEIGHT
FROM dw.FCLT_ROOMS AS fr
JOIN dw.BUILDINGS AS b ON fr.FCLT_BUILDING_KEY = b.BUILDING_KEY
JOIN dw.FCLT_BUILDING_ADDRESS AS fba 
  ON b.BUILDING_KEY = fba.FCLT_BUILDING_KEY
JOIN dw.COURSE_CATALOG_SUBJECT_OFFERED AS ccso 
  ON fr.FCLT_ROOM_KEY = ccso.MEET_PLACE
LEFT JOIN dw.FAC_BUILDING AS fb 
  ON b.BUILDING_KEY = fb.FAC_BUILDING_KEY
WHERE ccso.DEPARTMENT_NAME = 'Electrical Eng & Computer Sci'
\end{lstlisting}

{\footnotesize
\begin{tabular}{@{}llp{3.4cm}@{}}
\toprule
\textbf{Dimension} & \textbf{Score} & \textbf{Assessment} \\
\midrule
Executability & 2/2 & Executes successfully \\
Schema Validity & 2/2 & All tables exist and relevant \\
Filter Correctness & 2/2 & Department filter reasonable \\
Join Relevance & 2/2 & Six-table join path correct \\
Result Fidelity & 1/2 & Interpretation differs from gold \\
\midrule
\textbf{Total} & \textbf{8/10} & \\
\bottomrule
\end{tabular}
}

\textbf{Verdict:} \textit{partial} --- answers a reasonable interpretation.

\textbf{Issues:} Filters by department rather than student enrollment eligibility; gold uses different interpretation involving major requirements.
\end{partialbox}

\paragraph{Analysis.} This exemplifies why open-ended evaluation is necessary. The SQL scores 8/10 and produces useful results, but ``subjects CS students can enroll in'' admits multiple interpretations: (a) subjects offered by CS department, (b) subjects CS majors must take, (c) subjects with no restrictions. Traditional exact-match would mark this wrong despite practical utility.

\subsection{Low Quality: Missing Entity Coverage}

\begin{failurebox}{Example 4: Mailing List Instructor Query --- Score: 5/10}
\textbf{Question:} For each course instructor in the `keeper-zephyr' mailing list, provide mailing lists they subscribe to, instructor name, earliest/latest publication years, and total enrolled students.

\textbf{Question Type:} Deterministic

\textbf{Generated SQL:}
\begin{lstlisting}[basicstyle=\ttfamily\scriptsize]
SELECT NULL AS mailing_list_name, LCI.INSTRUCTOR_NAME,
  NULL AS earliest_publication_year,
  NULL AS latest_publication_year,
  SUM(SO.SUBJECT_ENROLLMENT_NUMBER) AS total_enrolled
FROM DW.LIBRARY_COURSE_INSTRUCTOR AS LCI
JOIN DW.LIBRARY_RESERVE_MATRL_DETAIL AS LRMD
  ON LCI.LIBRARY_COURSE_INSTRUCTOR_KEY = 
     LRMD.LIBRARY_COURSE_INSTRUCTOR_KEY
JOIN DW.SUBJECT_OFFERED AS SO ON LRMD.SUBJECT_ID = SO.SUBJECT_ID
WHERE LCI.DEPARTMENT = 'keeper-zephyr'
GROUP BY LCI.INSTRUCTOR_NAME
\end{lstlisting}

{\footnotesize
\begin{tabular}{@{}llp{3.4cm}@{}}
\toprule
\textbf{Dimension} & \textbf{Score} & \textbf{Assessment} \\
\midrule
Executability & 2/2 & Executes without errors \\
Schema Validity & 2/2 & Tables exist (but wrong ones) \\
Filter Correctness & 0/2 & Uses department, not mailing list \\
Join Relevance & 1/2 & Joins valid but miss MOIRA tables \\
Result Fidelity & 0/2 & Returns NULLs for key fields \\
\midrule
\textbf{Total} & \textbf{5/10} & \\
\bottomrule
\end{tabular}
}

\textbf{Verdict:} \textit{partial} --- executes but fails to answer.

\textbf{Issues:} Missing MOIRA\_LIST/MOIRA\_LIST\_MEMBER tables; incorrectly interprets `keeper-zephyr' as department filter; returns NULL for publication years.
\end{failurebox}

\paragraph{Analysis.} Entity extraction failed to connect ``mailing list'' to the MOIRA schema. The system recognized ``instructor'' and ``enrolled students'' but missed the link to mailing list membership. When questions span multiple domains (academic + mailing lists), entity coverage may be incomplete.

\subsection{Failure Pattern Summary}

\begin{table}[h]
\centering
\small
\begin{tabular}{lcc}
\toprule
\textbf{Failure Pattern} & \textbf{Count} & \textbf{\%} \\
\midrule
Missing entity coverage & 12 & 40\% \\
Disconnected graph regions & 10 & 33\% \\
Ambiguous terminology & 5 & 17\% \\
Schema name confusion & 3 & 10\% \\
\bottomrule
\end{tabular}
\caption{Failure pattern analysis on low-scoring BEAVER queries (n=30).}
\label{tab:failure_patterns}
\end{table}

\paragraph{Implications.} Entity coverage failures (40\%) suggest expanding entity extraction through iterative refinement would yield largest improvements. Disconnected graph regions (33\%) indicate implicit join discovery does not fully bridge all semantic relationships.

\section{Comparison with Schema-Aware Methods}
\label{app:dinsql}

To assess whether traditional text-to-SQL methods could work in our setting given reconstructed schema documentation, we evaluate DIN-SQL~\cite{pourreza2023dinsql} on our Amazon deployment. DIN-SQL decomposes text-to-SQL into schema linking, query classification, SQL generation, and self-correction modules. We convert our generated metadata to Spider format, providing DIN-SQL with the same table and column descriptions available to our system.

\paragraph{Setup.} We run DIN-SQL on all 240 tables using the converted schema. Evaluation uses blind LLM-as-judge pairwise comparison, identical to our main experiments.

\paragraph{Results.} Our system wins 72\% of comparisons, DIN-SQL wins 12\%, with 16\% ties. This gap persists despite DIN-SQL receiving complete schema documentation with human-readable descriptions. 

The limitation lies in schema linking: DIN-SQL must map question terms to schema elements for each query independently, without knowledge of which tables are commonly queried together or how domain terminology has historically mapped to schema elements. Our knowledge graph captures these patterns, accumulated from 500K historical queries, enabling more reliable retrieval when users employ business terminology that doesn't directly match schema names. This finding suggests that enterprise text-to-SQL benefits from retrieval grounded in historical usage patterns rather than per-query schema linking alone.

\end{document}